\documentclass[letterpaper, 10 pt, conference]{ieeeconf}  

\usepackage{amsmath}
\usepackage{graphicx}
\usepackage{float}
\usepackage{xcolor}
\usepackage{amssymb}
\usepackage{mathtools}
\usepackage{multirow}
\usepackage{caption}
\usepackage{subcaption}
\usepackage[export]{adjustbox}
\usepackage{soul}
\usepackage{makecell}
\usepackage{url}
\IEEEoverridecommandlockouts                              

\title{\LARGE \bf
PriorFormer: A Transformer for Real-time Monocular  \\ 3D Human Pose Estimation with Versatile Geometric Priors
}

\author{Mohamed Adjel$^{1,2}$ and Vincent Bonnet$^{1,4}$
\thanks{${^1}$ LAAS-CNRS, Université Paul Sabatier, CNRS, Toulouse, France.}
\thanks{${^2}$ NaturalPad, Montpellier, France. }
\thanks{${^3}$ CNRS-AIST JRL, IRL, AIST Tsukuba Headquarters and Information Technology Collaborative Research Center, Tsukuba, Japan.}
\thanks{${^4}$Image and Pervasive Access Laboratory (IPAL), CNRS-UMI, 2955, Singapore.
}}

\begin{document}

\maketitle
\thispagestyle{empty}
\pagestyle{empty}

\begin{abstract} 
This paper proposes a new lightweight Transformer-based lifter that maps short sequences of human 2D joint positions to 3D poses using a single camera. The proposed model takes as input geometric priors including segment lengths and camera intrinsics and is designed to operate in both calibrated and uncalibrated settings. To this end, a masking mechanism  enables the model to ignore missing priors during training and inference. This yields a single versatile network that can adapt to different deployment scenarios, from fully calibrated lab environments to in-the-wild monocular videos without calibration.

The model was trained using 3D keypoints from AMASS dataset with corresponding 2D synthetic data generated by sampling random camera poses and intrinsics.
It was then compared to an expert model trained, only on complete priors, and the validation was done by conducting an ablation study. Results show that both, camera and segment length priors, improve performance and that the versatile model outperforms the expert, even when all priors are available, and maintains high accuracy when priors are missing.  Overall the average 3D joint center positions estimation accuracy was as low as 36mm improving state of the art by half a centimeter and at a much lower computational cost. Indeed, the proposed model runs in $380\mu$s on GPU and $1800\mu$s on CPU, making it suitable for deployment on embedded platforms and low-power devices.

\end{abstract}

\section{INTRODUCTION}
\label{intro}

3D Human Pose Estimation (3D-HPE) is a foundational task in computer vision that aims to determine the 3D positions of human joints from visual data. Applications such as human-robot interaction \cite{hover}, action recognition \cite{avinash}, biomechanics \cite{opencap}, or humanoid robot teleoperation \cite{hover} require real-time performance, but often operate in settings where computational resources are limited. In such contexts, the ability to run lightweight pose estimation models on edge devices such as smartphones, embedded GPUs, or microcontrollers is of crucial importance.

Two main paradigms dominate monocular 3D pose estimation: direct image-to-3D regression \cite{pavlakos2016_3DHPE}, and two-stage pipelines that lift 2D keypoints to 3D \cite{einfalt2023_lifting}. The first approach maps RGB images directly to 3D joint coordinates using large neural networks trained on datasets such as Human3.6M \cite{h36m}, but suffers from high computational cost, limited dataset diversity, and annotation noise. In contrast, the second approach first estimates 2D keypoints using well established backbones such as HRNet \cite{hrnet} or RTMPose \cite{rtmpose}, and then lifts them to 3D using dedicated  architectures \cite{2dto3d_2017, 2dto3d_lifter_anatomical,  einfalt2023_lifting,Wen2024_semgan_lifting, 2Dto3D_2025}. The lifting strategy decouples geometry from vision, which enables training on large-scale MoCap datasets \cite{amass}, and significantly reduces computational cost, making it more suitable for edge devices deployment.

However, 2D to 3D lifting remains ill-posed without strong priors, and most accurate approaches require computationally heavy architectures \cite{Wen2024_semgan_lifting}. Camera intrinsics and subject-specific anthropometry are crucial to resolve depth ambiguities, yet most lifting models ignore these information due to the difficulty of acquiring their values outside of a laboratory setting. Recent advances in computer vision demonstrate that camera and anatomical priors could be accurately estimated from raw monocular video alone \cite{self_cam_calib_2022, cam_self_calib_2023, humanNerf2024, airnerf2024, cameraHMR_2024, blade2025}, enabling the use of subject- and environment-specific cues without requiring cumbersome calibration routines as in traditional biomechanical analysis \cite{opencap, pagnon2022_pose2sim, adjel_iros_2023}.

In this context, this paper proposes to train a lightweight Transformer-based lifter that maps short sequences of 2D joint positions to 3D poses. Our model takes as input optional geometric priors including bone lengths and camera intrinsics and is designed to operate in both calibrated and uncalibrated settings. To this end, we use a masking mechanism that enables the model to ignore missing priors during training and inference. This yields a single versatile network that can adapt to different deployment scenarios, from fully calibrated lab environments to in-the-wild uncalibrated monocular videos.



\section{RELATED WORK}
\label{related_works}

\subsection{Lifting using Transformers} 
Several works have adopted Transformer architectures for 2D-to-3D human pose lifting, leveraging their strength in modeling long-range dependencies across joint sequences. PoseFormer \cite{poseformer2021} introduced a spatio-temporal transformer design to capture intra-frame joint relations followed by inter-frame motion modeling. MHFormer \cite{mhformer2023} extended this architecture by generating multiple 3D hypotheses to account for depth ambiguity, improving robustness but increasing inference time and memory usage. JointFormer \cite{jointformer2022} proposed alternating spatial-temporal layers to refine joint-specific motion cues, yet still operates with dense attention and lacks explicit geometric conditioning. More recently, PoseFormerV2 \cite{poseformerv2_2023} addressed computational cost and robustness but still assumes long input sequences and does not handle missing geometric priors. In contrast we propose a tiny Transformer lifter that is magnitudes smaller than state-of-the-art architectures, uses short sequences of 2D keypoints, and optional geometric priors. Furthermore, the approaches assume fixed inputs while our model is trained to operate under varying prior availability, from fully calibrated setups to uncalibrated in-the-wild scenarios, which eliminates the need for multiple expert networks. 

\subsection{Lifting with geometric priors}
Numerous studies have explored the use of geometric priors to improve 2D-to-3D pose lifting. Akhter and Black \cite{akhter_black_2015_priors} proposed pose-conditioned joint angle limits to constrain the 3D solution space, using a sparse dictionary and binary joint validity priors. While effective, their method relies on handcrafted constraints and sparse coding, which are not suited for real-time inference. More recently, Nie et al. \cite{Nie2023_priors_lifting} introduced a latent 3D body concept shared between 2D and 3D domains via domain adaptation, using it as a learned prior for semi-supervised lifting. However, this approach requires complex training and does not expose interpretable priors such as camera or anatomical parameters. In contrast, Chen et al. \cite{Chen2021_priors} explicitly concatenate camera intrinsics and bone lengths as input to an MLP lifter, and show that both priors improve accuracy. Yet, they assume full prior availability, do not exploit temporal information, and do not generalize across camera setups. Our work will extend these by incorporating geometric priors in a learnable and interpretable way, while being robust to missing inputs via a masking mechanism. 


\section{METHOD}
\label{sec:method}
\begin{figure*}[t]
    \centering
    \includegraphics[width=1.0\linewidth]{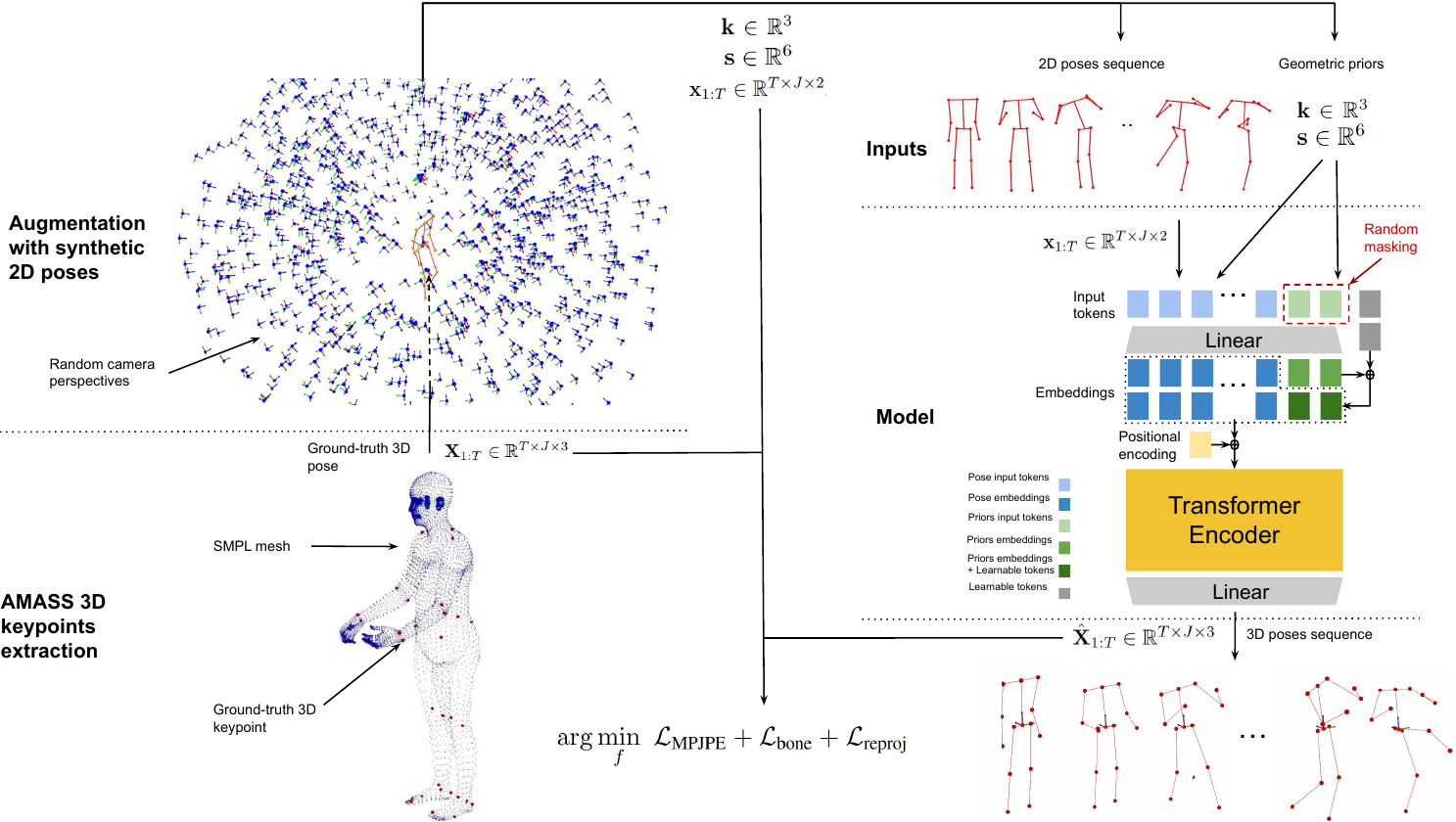}
    \caption{Overview of the training and inference pipeline. 
    3D joint center positions are extracted from SMPL \cite{smpl-x} meshes of the AMASS \cite{amass} dataset and used to generate synthetic 2D pose sequences by projecting them under randomly sampled camera intrinsics and viewpoints. 
    These sequences, along with geometric priors (camera intrinsics and segment lengths), are used to train a Transformer-based lifter.
    The model takes as input $T \times J$ pose tokens and two randomly masked prior tokens, all embedded in a shared latent space, and processes them via a lightweight Transformer encoder.
    Learnable prior tokens are summed with the projected (latent) priors and prepended to the sequence of pose embeddings.
    The output embeddings are decoded into 3D joint coordinates, supervised by ground-truth 3D poses and additional loss terms (bone lengths and reprojection consistency).}
    \label{fig:overview_figure}
\end{figure*}

Fig. \ref{fig:overview_figure} shows an overview of the proposed pipeline with data augmentation, Transformer model, geometric priors and training strategy.

The proposed model was trained using 3D keypoints were extracted from AMASS \cite{amass} dataset  and generated corresponding 2D projections by sampling random camera poses and intrinsics parameters. This data augmentation allowed us to expose the network to a large diversity of viewpoints and camera configurations, improving generalization across perspectives. The model was trained on joint center keypoints, but our pipeline is agnostic to the keypoint definition and can be extended to denser sets of anatomical landmarks, as proposed in \cite{opencapbench}. The full training and evaluation pipeline is reproducible, and the pre-trained versatile model is provided\footnote{https://github.com/mohamdev/2D-to-3D-pose-lifter}. 



\subsection{Problem Formulation}
\label{sec:problem}

We address the problem of lifting 2D joint sequences to their corresponding 3D poses in a pelvis-centered coordinate frame. Given a short sequence of 2D joint positions $\mathbf{x}_{1:T} \in \mathbb{R}^{T \times J \times 2}$ extracted from monocular video, the objective is to predict the corresponding 3D joint sequence $\mathbf{X}_{1:T} \in \mathbb{R}^{T \times J \times 3}$, where $T = 13$ is the temporal window size, chosen small for real-time inference, and $J = 12$ is the number of joint center keypoints.

The model also receives optional geometric priors. The first is a 3D camera intrinsics vector $\mathbf{k} \in \mathbb{R}^3$, consisting of a normalized focal length and principal point offset, which helps to account for perspective. The second is a vector of bone segment lengths $\mathbf{s} \in \mathbb{R}^6$, representing subject-specific limb proportions computed from six fixed joint pairs at the central frame of the sequence.

We aim to learn a function
\[
f(\mathbf{x}_{1:T}, \mathbf{k}, \mathbf{s}) \rightarrow \hat{\mathbf{X}}_{1:T},
\]
that maps a 2D joint sequence and available priors to 3D joint coordinates. At training and inference time, priors may be partially or fully masked, allowing the network to adapt to different levels of calibration and anatomical knowledge.

Each training example is synthesized by projecting 3D joint sequences from AMASS \cite{amass} into 2D, using randomly sampled camera intrinsics and extrinsics (see Figure \ref{fig:random_cameras}). This creates diverse input views and projection parameters across training. To increase model robustness, priors are stochastically masked during training. The model is trained to minimize a weighted combination of these three loss terms:
\[
\arg\min_f \ \mathcal{L}_{\text{MPJPE}} + \mathcal{L}_{\text{bone}} + \lambda_{\text{reproj}} \cdot \mathcal{L}_{\text{reproj}},
\]
with $\lambda_{\text{reproj}} = 10^{-2}$. All terms are averaged over the batch. The first term, $\mathcal{L}_{\text{MPJPE}}$, is the mean per-joint position error (MPJPE) between the predicted and ground-truth 3D joint coordinates. It provides direct supervision in 3D space. The second term, $\mathcal{L}_{\text{bone}}$, penalizes deviations in segment lengths for six selected bone pairs at the central frame, encouraging anatomically plausible outputs. The third term, $\mathcal{L}_{\text{reproj}}$, is a full-perspective reprojection loss computed in pixels, which compares the reprojected 3D joints with the input 2D observations using ground-truth camera intrinsics and extrinsics. The reprojection loss is scaled by the inverse of the image width and weighted by a small coefficient $\lambda_{\text{reproj}}$ to balance its contribution relative to the 3D space losses.

\subsection{Transformer-Based Lifter Architecture}
\label{sec:architecture}

\begin{figure}[t]
    \centering
    \includegraphics[width=0.9\linewidth]{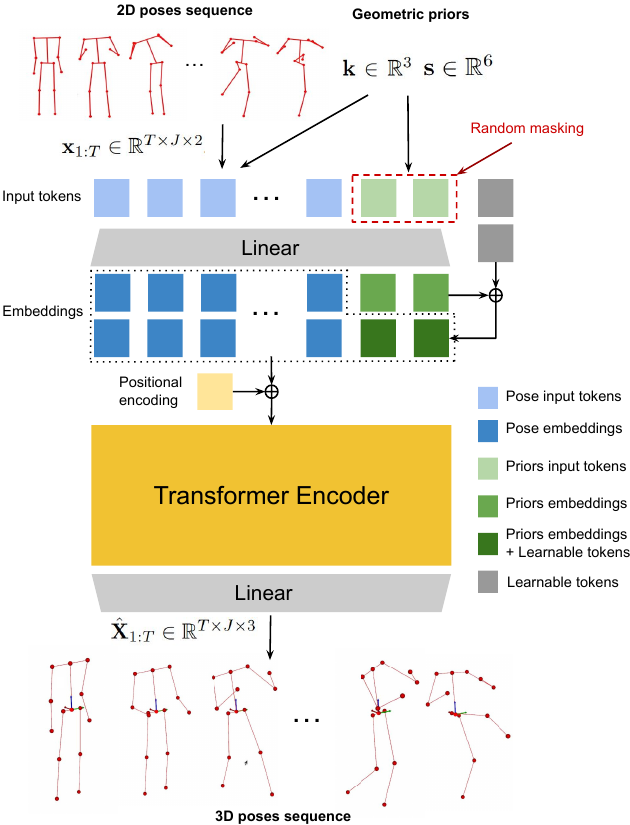}
    \caption{Overview of the Transformer-based lifter architecture. The input consists of $T \times J$ pose input tokens (2D keypoint coordinates concatenated with repeated geometric priors) and two additional prior input tokens: one for the camera intrinsics $\mathbf{k}$ and one for the segment lengths $\mathbf{s}$. Thes two tokens are randomly masked during the training of the versatile model.
    All tokens are first projected into a shared 128-dimensional embedding space.
    Prior tokens are enhanced by summing the embedding of the raw prior with a learned token.
    The full sequence (of length $T \cdot J + 2$) is then augmented with sinusoidal positional encodings and passed through a lightweight Transformer encoder.
    The resulting embeddings corresponding to pose tokens are decoded via a shared linear regressor to produce 3D joint coordinates.}
    \label{fig:transformer_model}
\end{figure}

Our lifting model is a simple and lightweight Transformer encoder that processes spatiotemporal sequences of 2D joint positions along with optional geometric priors (see Figure \ref{fig:transformer_model}). It takes as input a sequence of 2D keypoints $\mathbf{x}_{1:T} \in \mathbb{R}^{T \times J \times 2}$, a camera intrinsics vector $\mathbf{k} \in \mathbb{R}^3$, and a segment-length vector $\mathbf{s} \in \mathbb{R}^6$. The model outputs 3D joint predictions $\hat{\mathbf{X}}_{1:T} \in \mathbb{R}^{T \times J \times 3}$ in a pelvis-centered frame.

Each input pose token encodes a keypoint $(x, y)$ normalized coordinates, concatenated with the corresponding camera intrinsics $\mathbf{k}$ and segment lengths $\mathbf{s}$, which are repeated across all joints and frames. This results in $T \times J$ pose tokens, each of dimension $D_{\text{in}} = 11$, containing $2$ keypoint values, $3$ normalized intrinsics, and $6$ normalized bone lengths. These tokens are passed through a shared linear embedding layer that maps each 11-dimensional token to a vector of dimension $D = 128$.

In addition to these joint tokens, we introduce two special tokens for the geometric priors: one for the camera intrinsics $\mathbf{k}$ and one for the segment lengths $\mathbf{s}$ (see Figure \ref{fig:transformer_model}). Each is first projected into the embedding space via a separate linear layer and then summed with a learned token. These prior tokens are prepended to the embedded sequence of pose tokens, resulting an input sequence of length $T \cdot J + 2$, where the first two prepended tokens carry explicit global prior information.

This full sequence is then enhanced with fixed sinusoidal positional encodings and passed to a Transformer encoder composed of 4 layers, each using 4 attention heads and a feedforward dimension of 256. The output corresponding to the pose tokens is passed through a final linear regressor to produce the predicted 3D joint coordinates.

The architecture supports variable masking of priors at both training and inference (see Figure \ref{fig:transformer_model}). The final model contains under 0.6M parameters and runs in $380~\mu$s on an Nvidia RTX4070 GPU and $1800~\mu$s on an Intel Core i5-14400F CPU, making it suitable for real-time applications on relatively resource-constrained devices.

\subsection{Training with Prior Masking}
\label{sec:masking}

\begin{figure*}[t]
    \centering
    \includegraphics[width=\textwidth]{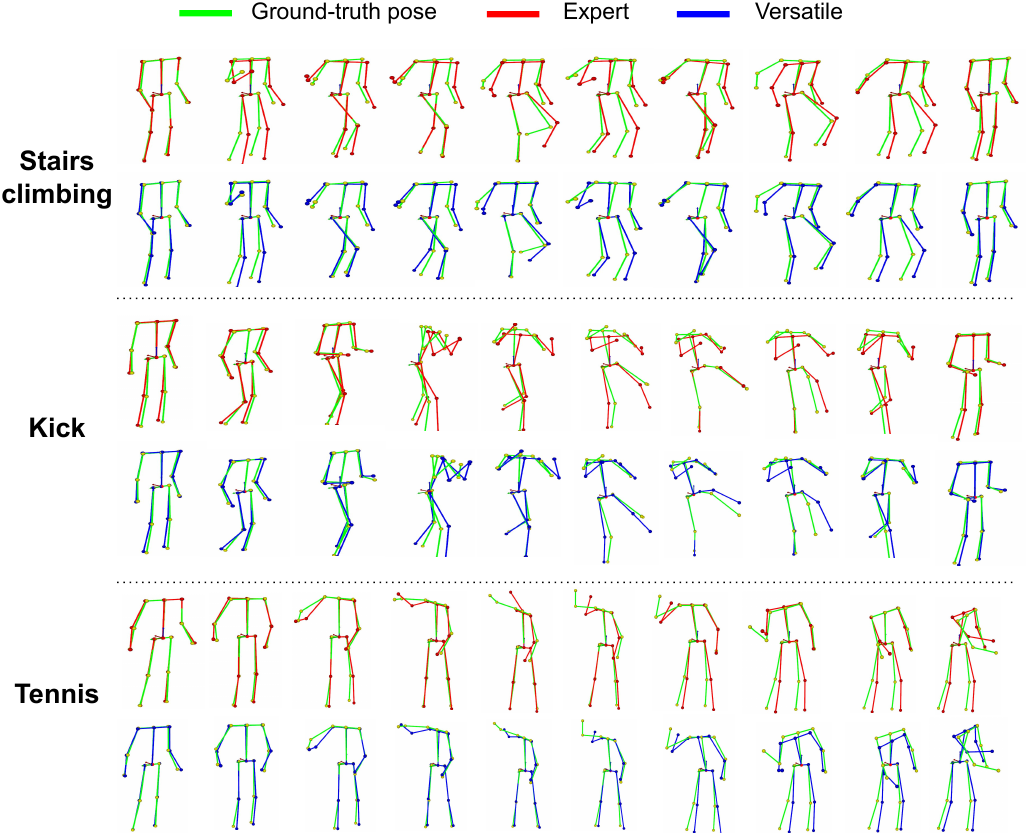}
    \caption{
    Qualitative comparison of 3D pose predictions on three representative motions (\textit{kick}, \textit{tennis}, \textit{stairs climbing}). 
    Ground-truth joint positions are shown in green, predictions from the expert model in red, and predictions from the versatile model in blue. All priors were available, for both versatile and expert model.}
    \label{fig:qualitative_motions}

\end{figure*}

We train two variants of the Transformer lifter: an expert model and a versatile model. See Figure \ref{fig:qualitative_motions} for examples of ground truth qualitative motions, predicted with both expert and versatile models. Both are initialized from the same pre-trained backbone trained for 50 epochs using a learning rate of $10^{-3}$ and a cosine learning rate scheduler. Fine-tuning is then performed for an additional 10 epochs with a reduced learning rate of $10^{-4}$, using the same optimizer and scheduler configuration. This ensures that both models see the same number of training steps, for a fair comparison.

The expert model is trained with complete access to all geometric priors throughout fine-tuning. In contrast, the versatile model is trained with stochastic masking of inputs to simulate varying deployment conditions (see Figure \ref{fig:transformer_model}). At each training step, priors are masked according to the following schedule: 25\% of the time, only camera intrinsics are masked; 25\% of the time, only segment lengths are masked; 25\% of the time, both are masked; and 25\% of the time, all priors are provided. This regular masking procedure encourages robustness to missing inputs and allows the model to generalize across calibrated and uncalibrated settings. When a prior is masked, its input vector is simply zeroed out. During fine-tuning, only the Transformer encoder and the final regression head are updated; all other parameters remain frozen. The model is trained using the AdamW optimizer with weight decay set to $10^{-4}$. We use the default dropout rate of 0.1 in each Transformer encoder layer, however, higher dropout would provide stronger regularization and may improve robustness.

\subsection{Dataset}
\label{sec:data_generation}

To train and evaluate our model, we construct a synthetic dataset derived from the AMASS motion capture repository~\cite{amass}. We select three representative subsets, ACCAD \cite{ACCAD_dataset}, BMLMovi \cite{Movi_dataset_2021}, and KIT \cite{KIT_dataset_2015}, which cover a wide range of motion styles and subjects. From these, we extract 13 human activities of interest (\textit{walk}, \textit{squat}, \textit{tennis}, \textit{bend}, \textit{displace/tilt}, \textit{wipe}, \textit{pour}, \textit{guitar}, \textit{stairs climbing}, \textit{push recovery}, \textit{punch}, \textit{turn}, \textit{kick}), totaling approximately 17 hours of motion data across 160 unique subjects.

\begin{figure}[t]
    \centering
    \includegraphics[width=0.9\linewidth]{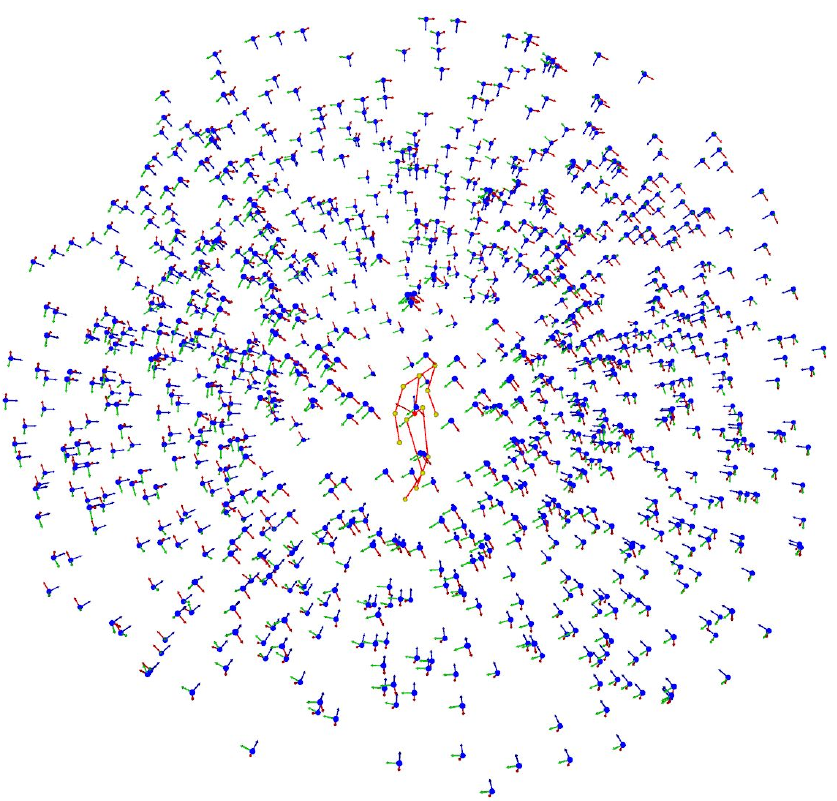}
    \caption{View of the camera perspectives randomly generated during learning. Each frame represents a unique camera pose, with randomized intrinsics parameters. During training, the ground-truth 3D keypoints $\mathbf{X}_{1:T} \in \mathbb{R}^{T \times J \times 3}$ (represented by the red skeleton) were projected into the randomly generated perspectives, and the corresponding 2D keypoints $\mathbf{x}_{1:T} \in \mathbb{R}^{T \times J \times 2}$ were used as input for the 2D-to-3D lifter.}
    \label{fig:random_cameras}
\end{figure}

We use SMPL \cite{smpl-x} meshes provided in AMASS to compute 3D joint center trajectories. We extract 12 anatomical keypoints (wrists, elbows, shoulders, hips, knees, ankles) as the barycenters of carefully chosen mesh vertex \cite{opencapbench} (see Figure \ref{fig:overview_figure}). Each pose sequence is centered in a pelvis-aligned root frame, built from the mid-hip position and a static orientation estimated from hip and shoulder keypoints. The resulting sequences of 3D joint positions are used as ground-truth labels.

\begin{figure*}[t]
    \centering
    \includegraphics[width=0.9\textwidth]{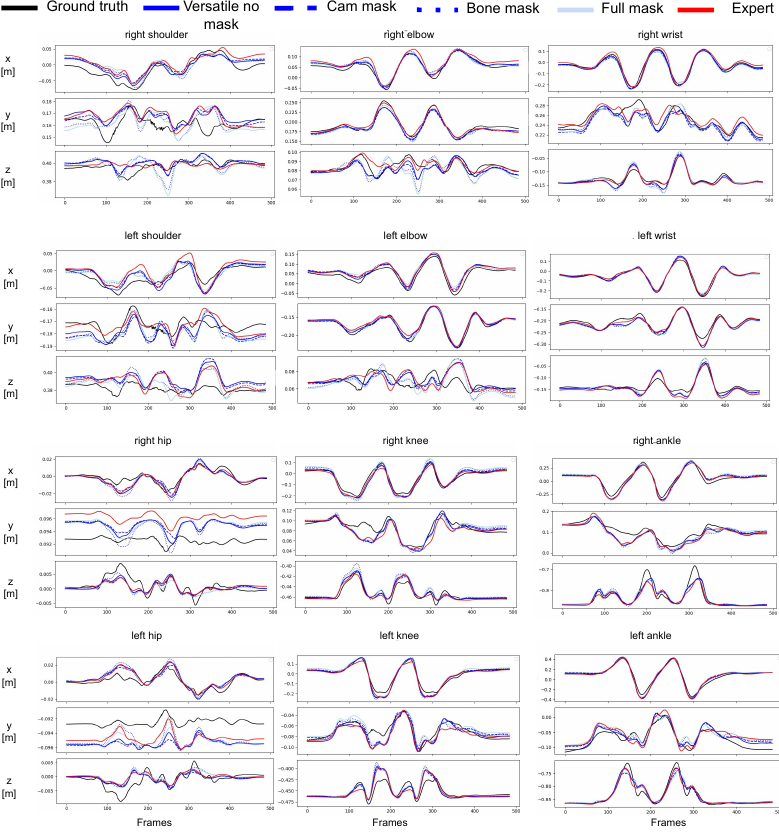}
    \caption{
    Comparison of 3D joint trajectories on a walking sequence.
    Each subplot shows the $x$, $y$, and $z$ coordinates over time for a single anatomical joint.
    Ground-truth trajectories are shown in black.
    The versatile model is evaluated under four masking modes: 
    no masking (solid blue), only camera intrinsics masked (dashed blue), only segment lengths masked (dotted blue), and both priors masked (light blue).
    Predictions from the expert model, trained with full priors only, are shown in red.
    The versatile model remains close to the ground truth across all masking conditions, demonstrating robustness to missing geometric information.
    }
    \label{fig:plot}
\end{figure*}

To generate input 2D observations, we apply random perspective projections to these 3D poses at training time. For each training sample, a synthetic monocular camera is randomly placed around the subject at a distance sampled uniformly between 1.5m and 3.5m, with its $z$-axis pointing towards mid-hip (see Figure \ref{fig:random_cameras}). Focal length and principal point offset are randomized to simulate different intrinsics. The resulting 2D projections are normalized and used as model input, along with the corresponding geometric priors (camera intrinsics and segment lengths), which are extracted from the original 3D data. This on-the-fly data generation exposes the model to a wide variety of camera viewpoints and configurations, while removing the need to store large volumes of projected data on disk, making the training pipeline lightweight and scalable. We randomly split the dataset into 90\% for training and 10\% for validation, ensuring subject and activity diversity in both sets.

\begin{table*}[ht]
\small\centering
\fontsize{8}{8}\selectfont
\caption{Per‐task MPJPE (mm) for the expert model under different ablation settings. 
\textit{Full priors}: no priors masked; 
\textit{Camera masked}: only camera priors masked; 
\textit{Bones masked}: only segment priors masked; 
\textit{No priors}: all priors masked.}
\label{tab:final_model_results}
\begin{tabular}{| l | c c c c |}
\hline
\textbf{Task} & \textbf{Full priors} & \textbf{Camera masked} & \textbf{Bones masked} & \textbf{No priors} \\
\hline
\hline
bend             & $37.1 \pm 18.3$ & $209.7 \pm 45.7$ & $397.8 \pm 12.9$ & $398.3 \pm 12.4$ \\
displace tilt   & $27.8 \pm 7.5$  & $202.6 \pm 51.0$ & $405.6 \pm 3.3$  & $403.2 \pm 3.1$ \\
guitar           & $27.0 \pm 7.2$  & $191.1 \pm 7.7$  & $440.2 \pm 0.3$  & $440.1 \pm 0.3$ \\
kick             & $46.8 \pm 13.9$ & $216.5 \pm 41.9$ & $426.0 \pm 16.0$ & $424.5 \pm 15.7$ \\
squat            & $50.5 \pm 2.0$  & $205.0 \pm 28.8$ & $472.7 \pm 3.4$  & $467.6 \pm 3.8$ \\
punch            & $56.1 \pm 19.4$ & $194.8 \pm 23.3$ & $448.2 \pm 11.5$ & $447.9 \pm 9.8$ \\
pour             & $36.8 \pm 4.6$  & $191.3 \pm 46.3$ & $406.2 \pm 13.3$ & $405.7 \pm 13.8$ \\
stairs climbing  & $34.3 \pm 9.6$  & $198.0 \pm 48.6$ & $407.2 \pm 8.7$  & $408.1 \pm 8.1$ \\
push recovery    & $27.4 \pm 9.0$  & $182.4 \pm 46.6$ & $400.1 \pm 18.2$ & $400.6 \pm 17.4$ \\
tennis           & $45.8 \pm 18.1$ & $226.6 \pm 36.8$ & $441.9 \pm 12.7$ & $441.8 \pm 12.1$ \\
turn             & $43.7 \pm 29.7$ & $229.1 \pm 55.7$ & $395.7 \pm 18.4$ & $396.7 \pm 18.2$ \\
walk             & $36.5 \pm 30.1$ & $199.2 \pm 53.4$ & $395.9 \pm 16.8$ & $396.5 \pm 16.7$ \\
wipe             & $45.3 \pm 26.9$ & $277.2 \pm 60.5$ & $423.4 \pm 32.7$ & $415.5 \pm 32.8$ \\
\hline
\textbf{Average} & $\mathbf{39.6 \pm 15.1}$ & $\mathbf{209.5 \pm 42.0}$ & $\mathbf{420.1 \pm 12.9}$ & $\mathbf{419.0 \pm 12.6}$ \\
\hline
\end{tabular}
\end{table*}

\begin{table*}[ht]
\small\centering
\fontsize{8}{8}\selectfont
\caption{Per‐task MPJPE (mm) for the fine-tuned versatile model under different ablation settings. 
\textit{Full priors}: no priors masked; 
\textit{Camera masked}: only camera priors masked; 
\textit{Bones masked}: only segment priors masked; 
\textit{No priors}: all priors masked.}
\label{tab:versatile_model_results}
\begin{tabular}{| l | c  c  c  c |}
\hline
\textbf{Task} & \textbf{Full priors} & \textbf{Camera masked} & \textbf{Bones masked} & \textbf{No priors} \\
\hline
\hline
bend             & $39.0 \pm 18.1$ & $46.4 \pm 29.5$ & $36.2 \pm 13.5$ & $43.1 \pm 17.6$ \\
displace tilt   & $23.1 \pm 4.6$ & $24.6 \pm 5.1$ & $23.7 \pm 3.8$ & $24.5 \pm 4.4$ \\
guitar           & $26.2 \pm 7.8$ & $18.1 \pm 1.7$ & $25.7 \pm 3.6$ & $23.8 \pm 6.2$ \\
kick             & $42.7 \pm 10.3$ & $48.6 \pm 13.1$ & $51.2 \pm 16.3$ & $55.1 \pm 15.9$ \\
squat            & $33.7 \pm 0.1$ & $46.2 \pm 0.5$ & $59.3 \pm 13.1$ & $52.9 \pm 11.8$ \\
punch            & $53.1 \pm 15.7$ & $56.0 \pm 14.4$ & $58.3 \pm 10.5$ & $66.0 \pm 22.5$ \\
pour             & $36.3 \pm 4.7$ & $36.6 \pm 1.6$ & $35.8 \pm 6.3$ & $38.1 \pm 7.2$ \\
stairs climbing & $33.7 \pm 8.4$ & $35.8 \pm 7.6$ & $38.8 \pm 16.1$ & $37.6 \pm 8.6$ \\
push recovery   & $27.1 \pm 9.0$ & $30.0 \pm 10.2$ & $28.2 \pm 8.6$ & $32.0 \pm 12.4$ \\
tennis           & $45.5 \pm 15.9$ & $44.1 \pm 5.1$ & $42.3 \pm 9.8$ & $44.4 \pm 10.1$ \\
turn             & $44.8 \pm 30.4$ & $47.9 \pm 40.8$ & $41.7 \pm 27.5$ & $45.5 \pm 33.4$ \\
walk             & $36.9 \pm 30.4$ & $38.5 \pm 17.7$ & $36.2 \pm 19.8$ & $42.7 \pm 27.4$ \\
wipe             & $30.7 \pm 10.6$ & $30.0 \pm 8.7$ & $32.5 \pm 17.4$ & $30.3 \pm 11.6$ \\
\hline
\textbf{Average} & $\mathbf{36.4 \pm 12.8}$ & $\mathbf{38.7 \pm 12.0}$ & $\mathbf{39.2 \pm 12.8}$ & $\mathbf{41.2 \pm 14.5}$ \\
\hline
\end{tabular}
\end{table*}

\section{RESULTS}
\label{validation}

Table~\ref{tab:final_model_results} and Table~\ref{tab:versatile_model_results} reports the per-task MPJPE of the expert and versatile models under different ablation settings. We consider four conditions: \textit{Full priors} (all priors available), \textit{Camera masked}, \textit{Bones masked}, and \textit{No priors}.

The expert model, trained only on fully specified inputs, achieves strong performance in the full-priors setting ($39.6 \pm 15.1$mm), but fails to generalize when priors are missing. Masking either bone lengths or camera intrinsics leads to a dramatic drop in accuracy, with MPJPE increasing beyond 200mm in all tasks, and up to 472.7mm in the worst case. This shows that the expert model is highly dependent on the availability of geometric priors, and does not generalize to uncalibrated conditions.

In contrast, the versatile model shows consistent and robust performance across all ablation settings. It achieves $36.4 \pm 12.8$mm with full priors, slightly outperforming the expert. This result is noteworthy, as the versatile model was trained with stochastic masking and never had access to guaranteed complete information. We attribute this gain to the regularization effect induced by the masking mechanism, which prevents overfitting and encourages the model to learn more generalizable representations.

When camera or bone priors are removed, the performance of the versatile model degrades slightly. Compared to the expert, it maintains much lower error in both the \textit{Camera masked} (38.7mm vs. 209.5mm) and \textit{Bones masked} (39.2mm vs. 420.1mm) conditions. This demonstrates its ability to dynamically adjust to the available information at inference time.

The \textit{No priors} setting further highlights the advantage of the versatile design: it achieves 41.2mm average MPJPE, nearly matching the full-prior accuracy of the expert model. Overall, we observe that bone priors contribute more to performance than camera intrinsics, but both types of priors are worth.

To quantify the importance of each prior, we compare the full-priors setting to the cases where one prior is masked (see Figure \ref{fig:plot}). On average, masking bone lengths degrades accuracy more than masking camera intrinsics. This suggests that anatomical information provides stronger disambiguation for monocular 3D reconstruction than camera calibration alone. However, combining both priors yields the best performance, confirming their complementary role.

In summary, the versatile model trained with random prior masking: outperforms the expert model even when all priors are available; maintains high accuracy across all ablation settings; demonstrates that geometric priors, especially bone lengths, significantly improve lifting accuracy; offers a flexible solution suitable for both calibrated and uncalibrated deployment.

\section{Discussion}
\label{sec:discussion}

This work presents a compact and versatile 2D-to-3D pose lifting model that supports geometric priors and operates in both calibrated and uncalibrated settings. While our approach demonstrates strong performance across a diverse set of motion types and subjects, it still can be improved and further validated.

First, the proposed model is evaluated on synthetic data derived from the AMASS dataset, using perfect 2D projections and noise-free priors. This provides precise ground-truth supervision and enables controlled ablation studies, but it does not fully reflect the challenges encountered in real-world settings. In particular, 2D keypoints obtained from images or video are subject to detection noise and occlusion \cite{vitpose, rtmpose}. Furhtermore, geometric priors, when estimated in real-world scenarios, are likely to be imprecise. Future work should incorporate noise perturbations during training and fine-tune/evaluate the model on datasets like Human3.6M \cite{h36m}, which feature real camera setups.

Second, our method uses a reduced skeleton of 12 anatomical joints, selected for their relevance in biomechanics and human robot interaction. In contrast, most recent lifting benchmarks operate on denser keypoint formats (e.g., COCO). Although this limits direct comparison with existing methods, our pipeline is designed to be agnostic to the keypoint format and could be extended to support denser skeletons. For example, it could be combined with 2D-HPE models trained on synthetic anatomical datasets such as SynthPose \cite{opencapbench}, providing a calibration-free alternative to methods like OpenCap \cite{opencap}, which require multi-camera setups to augment sparse keypoints with triangulated 3D information.

With under 0.6M parameters, the current model already achieves strong lifting accuracy and fast inference. In comparison to state-of-the-art architectures that use more than 8M parameters \cite{poseformer2021, Wen2024_semgan_lifting}, our model is much smaller and keeps low MPJPE comparable to heavier state-of-the-art \cite{poseformer2021, poseformerv2_2023, Wen2024_semgan_lifting} methods that show errors higher than $40$mm on Human3.6M benchmark. While our architecture is intentionally kept minimal for this study, its size leaves a margin for further capacity scaling while keeping real-time inference. 

An important observation is that the versatile model performs better on validation data than expert model which exhibit signs of overfitting. While expert performs slightly better on the training set ($36.5$mm MPJPE) than the versatile model ($39.4$mm), it presents higher errors on the validation set (see Tables \ref{tab:final_model_results}, \ref{tab:versatile_model_results}), and fails to generalize when priors are missing, leading to substantial performance drops. In contrast, the versatile model maintains strong validation accuracy, which suggests that random input masking during training encourages the model to cope with uncertainty and improves generalization.

Lastly, our results confirm that geometric priors help resolve depth ambiguities, with bone lengths being especially effective. Furthermore, our versatile training strategy shows that the model keeps a convenient MPJPE of 41.2mm even when all priors are masked. This pre-trained model can be fine-tuned on real-world datasets, and keep high accuracy in both laboratory and in-the-wild conditions, while being compatible for use on edge devices.

\section{Conclusion}
\label{sec:conclusion}

We introduced a lightweight Transformer-based lifter that maps short 2D joint sequences to 3D poses, with support for optional geometric priors. The model is trained using a versatile masking strategy that enables robust performance across calibrated and uncalibrated scenarios. Our results show that this approach keeps good performance even when priors are missing, and outperforms an expert model trained with full supervision.

Despite using a compact architecture with under 0.6M parameters, our model achieves low MPJPE across diverse motions, and runs in real time on both CPU and GPU. This makes it well-suited for deployment in embedded systems and low-power applications. While current results are based on synthetic projections, the proposed framework offers a strong baseline for future work, with a pre-trained model that can be fine-tuned on real-world datasets, noisy inputs, and denser anatomical keypoints.

\section*{ACKNOWLEDGEMENT}
Blank for double-anonymous review process

\bibliographystyle{IEEEtran}

\bibliography{figarohiros2023.bib}

\end{document}